\pdfoutput=1

\documentclass[11pt]{article}

\usepackage{EMNLP2022}
\usepackage{times}
\usepackage{latexsym}
\usepackage{amsmath}
\usepackage{amssymb}
\usepackage{enumitem}
\usepackage{mathtools}
\usepackage{multirow}
\usepackage{booktabs}
\usepackage{multicol}
\usepackage{multirow}
\newtheorem{theorem}{Theorem}
\usepackage[T1]{fontenc}

\usepackage[utf8]{inputenc}
\usepackage{times}
\usepackage{latexsym}

\usepackage[T1]{fontenc}

\usepackage[utf8]{inputenc}

\usepackage{microtype}

\usepackage{inconsolata}

\title{Adaptive Contrastive Learning on Multimodal Transformer for \\ Review Helpfulness Predictions}

\author{Thong Nguyen$^{1, 2}$,~~Xiaobao Wu$^{3}$,~~Anh Tuan Luu$^{3}$\thanks{~~Corresponding Author}, \\
~~\textbf{Cong-Duy Nguyen}$^{3}$, \textbf{Zhen Hai}$^{4}$,~~\textbf{Lidong Bing}$^{4}$ \\
  $^1$National University of Singapore, Singapore \\
  $^2$VinAI Research, Vietnam \\
  $^3$Nanyang Technological University, Singapore \\
  $^4$DAMO Academy, Alibaba Group\\
  \texttt{\small thong.nguyen@u.nus.edu, anhtuan.luu@ntu.edu.sg} \\}

\begin{document}
\maketitle
\begin{abstract}
Modern Review Helpfulness Prediction systems are dependent upon multiple modalities, typically texts and images. Unfortunately, those contemporary approaches pay scarce attention to polish representations of cross-modal relations and tend to suffer from inferior optimization. This might cause harm to model’s predictions in numerous cases. To overcome the aforementioned issues, we propose Multi-modal Contrastive Learning for Multimodal Review Helpfulness Prediction (MRHP) problem, concentrating on mutual information between input modalities to explicitly elaborate cross-modal relations. In addition, we introduce Adaptive Weighting scheme for our contrastive learning approach in order to increase flexibility in optimization. Lastly, we propose Multimodal Interaction module to address the unalignment nature of multimodal data, thereby assisting the model in producing more reasonable multimodal representations. Experimental results show that our method outperforms prior baselines and achieves state-of-the-art results on two publicly available benchmark datasets for MRHP problem. 
\end{abstract}
\section{Introduction}
Current e-commerce sites such as Amazon, Ebay, etc., construct review platforms to collect user feedback concerning their products. These platforms play a fundamental role in online transactions since they help future consumers collect useful reviews which assist them in deciding whether to make the purchase or not. Unfortunately, nowadays the number of user-generated reviews is overwhelming, raising doubts related to the relevance and veracity of reviews. Therefore, there is a need to verify the quality of reviews before publishing them to prospective customers. As a result, this inspires a recent surge of interest targeting the Review Helpfulness Prediction (RHP) problem.
\begin{table}[h!]
\centering
\begin{tabular}{p{0.9\linewidth}}
\textbf{Product Information} \\
\small{
The Cooks Standard 6-Quart Stainless Steel Stockpot with Lid is made with 18/10 stainless steel with an aluminum disc layered in the bottom. The aluminum disc bottom provides even heat distribution and prevents hot spots. Tempered glass lid with steam hole vent makes viewing food easy. Stainless steel riveted handles offer durability. Induction compatible. Works on gas, electric, glass, ceramic, etc. Oven safe to 500F, glass lid to 350F. Dishwasher safe.} \\
\includegraphics[width=0.3\linewidth]{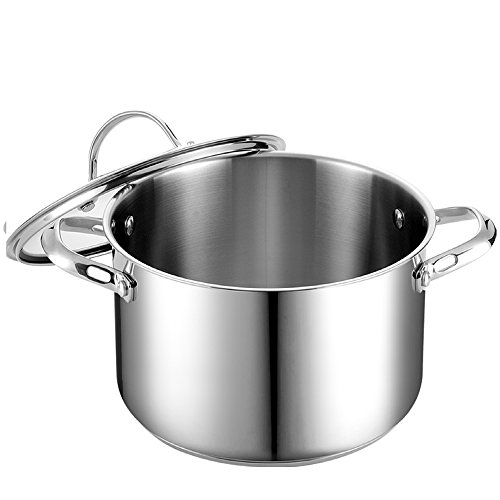} \includegraphics[width=0.3\linewidth]{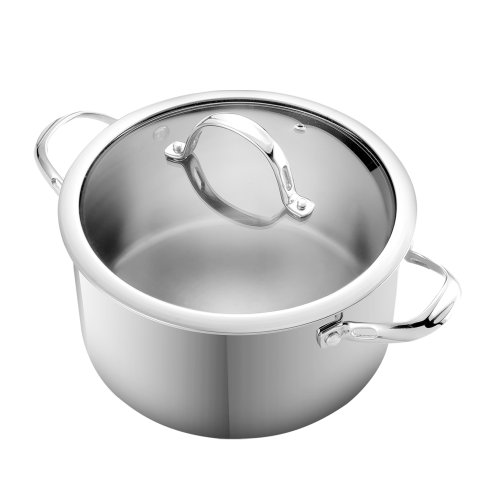} \\
\midrule
\textbf{Review 1} \\
\small{
I needed a stainless steel pot for canning my tomatoes.  I learned the hard way that you have to use a non-reactive pot or else your end result will be inedible (I thought I was using stainless steel but quickly realized it wasnt)  I headed to Amazon and came across this Cooks Standard SS Cookpot with cover and bought it after reading the reviews.  I have had it for just under a year and it still looks just as good as the day I bought it.  I couldn't be happier with my purchase!  Oh, and by the way, this one actually is stainless steel unlike the other pot I bought that said it was and wasn't.} \\
\midrule
\textbf{Review 2} \\
\small{
I ordered it on May 21st. What a waste of time and money.} \\
\includegraphics[width=0.3\linewidth]{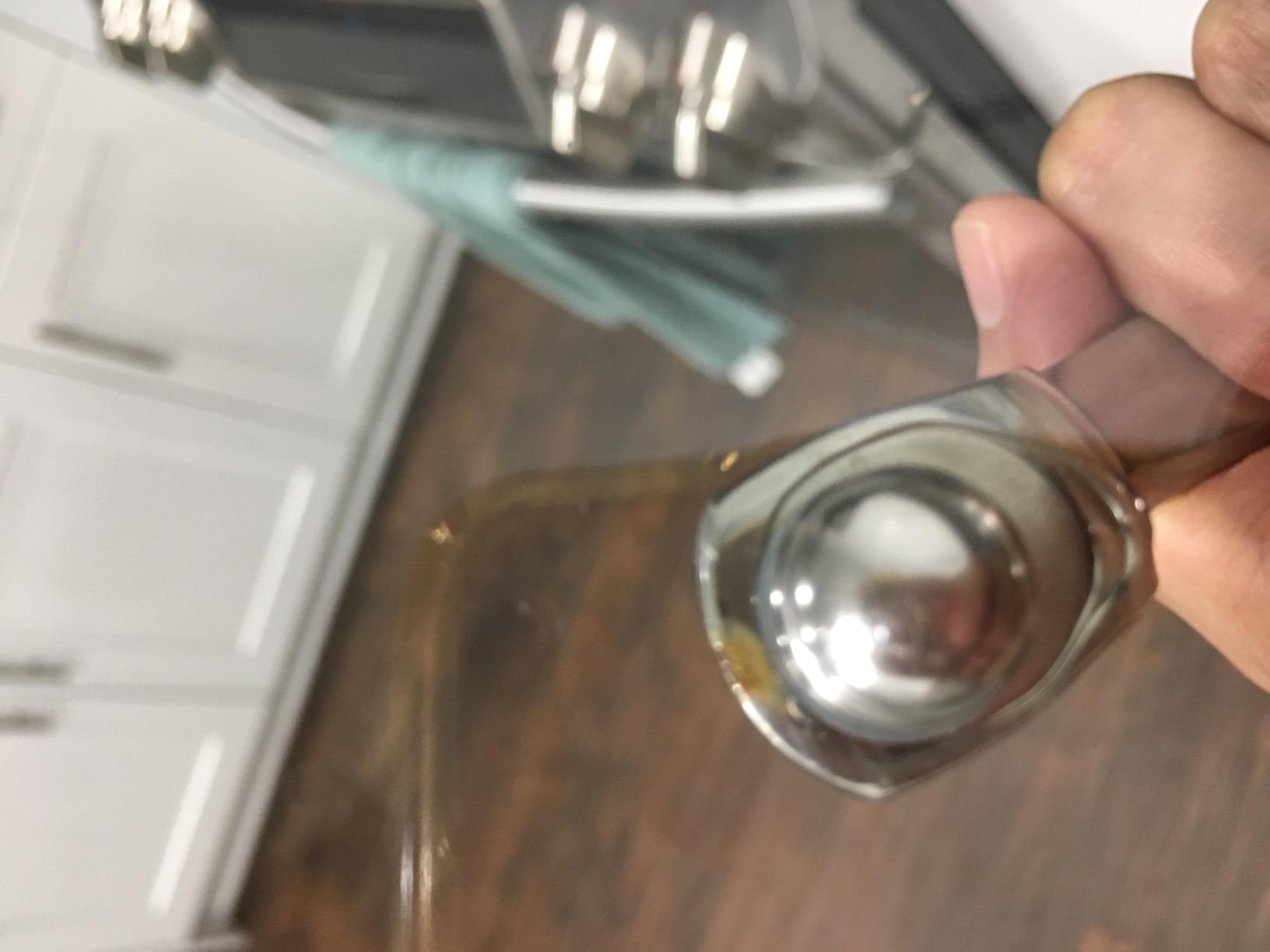} \\
\end{tabular}
\small
\begin{tabular}{ccc}
\toprule
& Review 1 & Review 2 \\
\midrule
Label score & 4 & 1 \\
MCR score & 0.168 & 3.637 \\
Our Model score & \textbf{4.651} & \textbf{0.743} \\
\bottomrule
\end{tabular}
\caption{Example of unreasonable predictions in the Multimodal Review Helpfulness Prediction task.}
\label{table:example}
\vspace{-5mm}
\end{table}

Two principal groups of early efforts focus on purely textual data. The first group follows feature engineering techniques, retrieving argument-based features \cite{liu2017using}, lexical features \cite{krishnamoorthy2015linguistic}, and semantic features \cite{kim2006automatically}, as input to their classifier. Inherently, their methods are labor-intensive and vulnerable to the typical issues of conventional machine learning methods. Instead of relying on manual features, the second group leverages deep neural models, for instance, RNN \cite{alsmadi2020predicting} and CNN \cite{chen2018cross}, to learn rich features automatically. Nonetheless, their approach is ineffective because the helpfulness of a review is not only contingent upon textual information but also other modalities.

To cope with the above issues, recent works \cite{liu2021multi,han2022sancl} proposed to utilize multi-modality via the Multi-perspective Coherent Reasoning (MCR) model. Hypothesizing that a review is helpful if it exhibits coherent text and images with the product information, those works take into account both textual and visual modality of the inputs, then estimate their coherence level to discern whether the reviews are \emph{helpful} or \emph{unhelpful}. However, the MCR model contains a detrimental drawback. Particularly, it aims to maximize the scores $s_p$ of positive (helpful) product-review pairs while minimizing those $s_n$ of negative (unhelpful) pairs. Hence, it was assumed that following the aforementioned manner would project features with similar semantics to stay close and those with disparate ones to be distant apart. Unfortunately, in multimodal learning, this was shown not to be the case, causing the model to learn ad-hoc representations \cite{zolfaghari2021crossclr}. This is one reason leading to unreasonable predictions of MCR in Table \ref{table:example}. As it can be seen, even though Review 1 closely relates to the product of ``\emph{6-Quart Stainless Steel Stockpot}'', the model classifies it as \emph{unhelpful}. In addition, the target of Review 2’s text content is vague because it does not specifically correspond to the ``\emph{Stockpot}''. In fact, it can be used for any product. Moreover, the image does not clearly show any hint of the ``\emph{Stockpot}'' as well. Despite such vagueness, the output of MCR for Review 2 is still \emph{helpful}. 

As a remedy to this problem, we propose Cross-modal Contrastive Learning to mine the mutual information of cross-modal relations in the input to capture more sensible representations. Nonetheless, plainly applying symmetric gradient pattern, which is similar to MCR that they assign equivalent penalty to $s_n$ and $s_p$, is inflexible. In cases that $s_p$ is small and $s_n$ is already negatively skewed, or both $s_p$ and $s_n$ are positively skewed, it is irrational to assign equivalent penalties to both $s_p$ and $s_n$. Last but not least, MCR directly leverages Coherent Reasoning, repeatedly enforcing alignment among modalities in the input. This ignores the unaligned nature of multimodal input, for example, images might only refer to a particular section in the text, hence do not completely align with the textual content. In consequence, strictly forming alignment can make the model learn inefficient multimodal representations  \cite{tsai2019multimodal}.

To overcome the above problems, we propose an adaptive scheme to accomplish the flexibility in the optimization of our contrastive learning stage. Finally, we propose to adopt a multimodal attention module that reinforces one modality’s high-level features with low-level ones of other modalities. This not only relaxes the alignment assumption but also informs one modality of information of others, encouraging refined representation learning.

In sum, our contributions are three-fold:
\begin{itemize}
    \item We propose an Adaptive Cross-modal Contrastive Learning for Review Helpfulness Prediction task by polishing cross-modal relation representations.
    \item We propose a Multimodal Interaction module which correlates modalities’ features without depending upon the alignment assumption.
    \item We conducted extensive experiments on two datasets for the RHP problem and found that our method outperforms other baselines which are both textual-only and multimodal, and obtains state-of-the-art results on those benchmarks.
\end{itemize}
\section{Model Architecture}
\begin{figure*}[t]
    \centering
    \includegraphics[width=\linewidth]{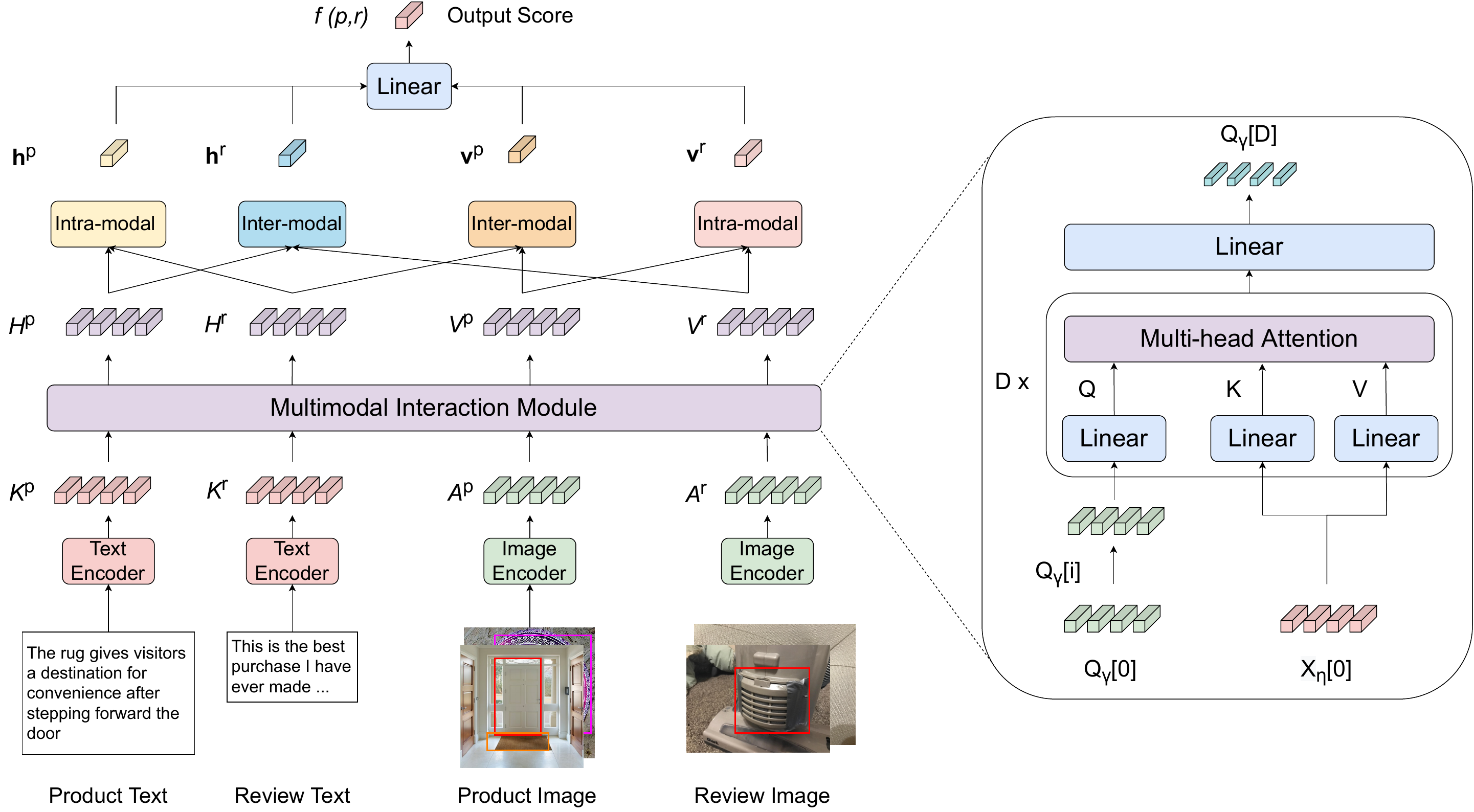}
    \caption{Diagram of our Multimodal Review Helpfulness Prediction model.}
    \label{fig:model}
\end{figure*}

In this section we delineate the overall architecture of our MRHP model. Particular modules of our system are depicted in Figure \ref{fig:model}.
\subsection{Problem Definition}
Given a product item $p$, which consists of a description $T^p$ and images $I^p$, and a set of reviews $R = \{r_1,…, r_N\}$, where each review is composed of user-generated text $T^r_i$ and images $I^r_i$, RHP model’s task is to generate the scores
\begin{equation}
    s_i = f(p, r_i), \quad 1 \leq i \leq N
\end{equation}
where $N$ is the number of reviews for product $p$ and $f$ is the scoring function of the RHP model. Empirically, each score estimated by $f$ indicates the helpfulness level of each review, and the ground-truth is the descending sort order of helpfulness scores.
\subsection{Encoding Modules}
Our model accepts product description $T^p$, product images $I^p$, review text $T^r_i$, and review images $I^r_i$ as input. The encoding process of those elements is described as follows.

\noindent\textbf{Text Encoding} Product description and review text are sequences of words. Each sequence is indexed into the word embedding layer and then passed into the respective LSTM layer for product or review.
\begin{gather}
    K^p = \text{LSTM}^p (\mathbf{W}_{\textbf{emb}} (T^p)) \\
    K^r = \text{LSTM}^r (\mathbf{W}_{\textbf{emb}} (T^r)) 
\end{gather}
where $K^p \in \mathbb{R}^{l_p \times d}$, $K^r \in \mathbb{R}^{l_r \times d}$, $l_p$ and $l_r$ are the sequence lengths of product and review text respectively, and $d$ is the hidden size.

\noindent\textbf{Image Encoding} We follow \citet{anderson2018bottom} to take detected objects as embeddings of the image. In particular, a pre-trained Faster R-CNN is applied to extract ROI features for $m$ objects $\{\mathbf{a}_1, \mathbf{a}_2, …, \mathbf{a}_m\}$ from the product and review images. Subsequently, we encode extracted features using the self-attention module (SelfAttn) \cite{vaswani2017attention}
\begin{equation}
    A = \text{SelfAttn}(\{\mathbf{a}_1, \mathbf{a}_2, ..., \mathbf{a}_m\})
\end{equation}
where $A \in \mathbb{R}^{m \times d}$ and $d$ is the hidden size. Here we use $A^p$ and $A^r$ to indicate product and review image features, respectively.

\subsection{Multimodal Interaction Module}
We consider two components $\gamma$, $\eta$ with their inputs $X_\gamma$, $X_\eta$, where $\eta$ is the concatenation of input elements apart from the one in $\gamma$. For instance, if $\gamma = {K^p}$, then $\eta = [K^r, A^p, A^r]$, where $[., .]$ indicates the concatenation operation. We define each cross-modal attention block to have three components $Q$, $K$, and $V$:
\begin{gather}
    Q_\gamma = X_\gamma \cdot W_{Q_\gamma} \\
    K_\eta = X_\eta \cdot W_{K_\eta} \\
    V_\eta = X_\eta \cdot W_{V_\eta}
\end{gather}
where $W_{Q_\gamma} \in \mathbb{R}^{d_\gamma \times d_k}$, $W_{K_\eta} \in \mathbb{R}^{d_\eta \times d_k}$, and $W_{V_\eta} \in \mathbb{R}^{d_\eta \times d_v}$ are weight matrices. The interaction between $\gamma$ and $\eta$ is computed in the cross-attention manner
\begin{equation}
    \begin{split}
        Z_\gamma = \text{CM}_{\gamma} (X_\gamma, X_\eta)
        = \text{softmax} \left(\frac{Q_\gamma \cdot K^T_{\eta}}{\sqrt{d_k}}\right) \cdot V_\eta
    \end{split}
\end{equation}
Our full module comprises $D$ layers of the above-mentioned attention block, as indicated in the right part of Figure \ref{fig:model}. Theoretically, the computation is carried out as follows
\begin{gather}
    Q_{\gamma}[0] = X_\gamma\\
    T[i] = \text{CM}_{\gamma}[i] (\text{LN}(Q_{\gamma}[i-1]), \text{LN}(X_\eta)) \\
    U_{\gamma}[i] = T[i] + Q_{\gamma}[i-1] \\
    Q_{\gamma}[i] = \text{GeLU}(\text{Linear}(U_{\gamma}[i])) 
\end{gather}
where $\textit{LN}$ denotes layer normalization operator. We iteratively estimate cross-modal features for product text, product images, review text, and review images with a view to obtaining $H^p$, $V^p$, $H^r$, and $V^r$.
\begin{gather}
    H^p = Q^p_k[D], \quad V^p = Q^p_a[D] \\
    H^r = Q^r_k[D], \quad V^r = Q^r_a[D] 
\end{gather}

After our cross-modal interaction module, we proceed to pass features to undertake relation fusion in three paths: intra-modal, inter-modal, and intra-review.

\noindent\textbf{Intra-modal Fusion} The intra-modal alignment is calculated for two relation kinds: (1) product text - review text and (2) product image - review image. Firstly, we learn alignment among intra-modal features via self-attention modules
\begin{gather}
    H^\text{intraM} = \text{SelfAttn}([H^p, H^r]) \\
    V^\text{intraM} = \text{SelfAttn}([V^p, V^r])
\end{gather}
Then intra-modal hidden representations are fed to a CNN, and continuously a max-pooling layer to attain salient entries
\begin{equation}
    \mathbf{z}^\text{intraM} = \text{MaxPool} (\text{CNN}([H^{\text{intraM}}, V^{\text{intraM}}]))
\end{equation}
\noindent\textbf{Inter-modal Fusion} Similar to intra-modal alignment, inter-modal one is calculated for two types of relations as well: (1) product text - review image and (2) product image - review text. The first step is also to relate feature components using self-attention modules
\begin{gather}
    H^{\text{prd\_txt - rvw\_img}} = \text{SelfAttn}([H^p, V^r]) \\
    H^{\text{prd\_img - rvw\_txt}} = \text{SelfAttn}([V^p, H^r])
\end{gather}
We adopt a mean-pool layer to aggregate inter-modal features and then concatenate the pooled vectors to construct the final inter-modal representation
\begin{gather}
    I^{\text{prd\_txt - rev\_img}} = \text{MeanPool}(H^{\text{prd\_txt - rvw\_img}})\\
    I^{\text{prd\_img - rev\_txt}} = \text{MeanPool}(H^{\text{prd\_img - rvw\_txt}})\\
    \mathbf{z}^{\text{interM}} = [I^{\text{prd\_txt - rvw\_img}}, I^{\text{prd\_img - rvw\_txt}}]
\end{gather}
\noindent\textbf{Intra-review Fusion} The estimation of intra-review module completely mimics the inter-modal manner. The only discrimination is that the estimation is taken upon two different relations: (1) product text - product image and (2) review text - review image.
\begin{gather}
    H^{\text{prd\_txt - prd\_img}} = \text{SelfAttn}([H^p, V^p]) \\
    H^{\text{rvw\_txt - rev\_img}} = \text{SelfAttn}([H^r, V^r]) \\
    G^{\text{prd\_txt - prd\_img}} = \text{MeanPool}(H^{\text{prd\_txt - prd\_img}}) \\
    G^{\text{rvw\_txt - rvw\_img}} = \text{MeanPool}(H^{\text{rvw\_txt - rvw\_img}}) \\
    \mathbf{z}^{\text{intraR}} = [G^{\text{prd\_txt - prd\_img}}, G^{\text{rvw\_txt - rvw\_img}}]
\end{gather}
Finally, we concatenate intra-modal, inter-modal, and intra-review output, and then feed the concatenated vector to the linear layer to obtain the ranking score:
\begin{gather}
\mathbf{z}^{\text{final}} = [\mathbf{z}^{\text{intraM}}, \mathbf{z}^{\text{interM}}, \mathbf{z}^{\text{intraR}}] \\
f(p,r_i) = \text{Linear}(\mathbf{z}^{\text{final}})
\end{gather}
\section{Training Strategies}
\subsection{Adaptive Cross-modal Contrastive Learning}
In this section, we explain the formulation and adaptive pattern along with its derivation of our Cross-modal Contrastive Learning.

\noindent\textbf{Cross-modal Contrastive Learning} First of all, we extract hidden states of helpful product-review pairs. Second of all, hidden features are max-pooled to extract meaningful entries.
\begin{gather}
    \mathbf{h}^p = \text{MaxPool}(H^p), \, \mathbf{h}^r = \text{MaxPool}(H^r) \\
    \mathbf{v}^p = \text{MaxPool}(V^p), \, \mathbf{v}^r = \text{MaxPool}(V^r) 
\end{gather}
We formulate our contrastive learning framework taking positive and negative pairs from the above-mentioned cross-modal features. In our framework, we hypothesize that pairs established by modalities of the same sample are positive, whereas those formed by modalities of distinct ones are negative. 
\begin{equation}
    \mathcal{L}_{\text{CE}} = -\sum_{i=1}^{B} \text{sim}(\mathbf{t}^1_i, \mathbf{t}^2_i) +  \sum_{j=1, k=1, j \neq k}^{B} \text{sim}(\mathbf{t}_j^{1}, \mathbf{t}_k^{2})
\end{equation}
where $\mathbf{t}^1, \mathbf{t}^2 \in \{\mathbf{h}^p, \mathbf{h}^r, \mathbf{v}^p, \mathbf{v}^r\}$, and $B$ denotes the batch size in the training process.

\noindent\textbf{Adaptive Weighting} The standard contrastive objective suffers from inflexible optimization due to irrational gradient assignment to positive and negative pairs. As a result, to tackle the problem, we propose the Adaptive Weighting Strategy for our contrastive framework. Initially, we introduce weights $\epsilon^p$ and $\epsilon^n$ to represent distances from the optimum, then integrate them into positive and negative terms of our loss.
\begin{equation}
\begin{split}
    &\mathcal{L}_{\text{AdaptiveCE}} = -\sum_{i=1}^{B} \epsilon^p_i \cdot \text{sim}(\mathbf{t}^1_i, \mathbf{t}^2_i) \\
    &+ \sum_{j=1, k=1, j \neq k}^{B} \epsilon_{j,k}^n \cdot \text{sim}(\mathbf{t}_j^{1}, \mathbf{t}_k^{2})
\end{split}
\label{eq:adaptive_ce}
\end{equation}
where $\epsilon_i^p = [o^p - \text{sim}(\mathbf{t}^1_i, \mathbf{t}^2_i)]_+$ and  $\epsilon_{j,k}^n = [\text{sim}(\mathbf{t}^1_j, \mathbf{t}^2_k) - o^n]_+$. Investigating the intuition to determine the values for $o^p$ and $o^n$, we continue to conduct derivation and arrive in the following theorem

\begin{theorem} Adaptive Contrastive Loss (\ref{eq:adaptive_ce}) has the hyperspherical form: 
\begin{equation*}
    \begin{split}
       &\mathcal{L}_{\text{AdaptiveCE}} = \sum_{i=1}^{B} \left(\text{sim} (\mathbf{t}^1_i, \mathbf{t}^2_i) - \frac{o^p}{2}\right)^2 \\
       &+ \sum_{j=1, k=1, j \neq k}^{B} \left(\text{sim} (\mathbf{t}^1_j, \mathbf{t}^2_k) - \frac{o^n}{2}\right)^2 - C, \\
       &\quad \text{where} \, C > 0         
    \end{split}
\end{equation*}
\label{theorem:spherical}
\end{theorem}
We provide the proof for Theorem (\ref{theorem:spherical}) in the Appendix section. As a consequence, theoretically the contrastive objective arrives in the optimum when $\text{sim}(\mathbf{t}_i^1, \mathbf{t}_i^2) = \frac{o^p}{2}$ and $\text{sim}(\mathbf{t}_j^1, \mathbf{t}_k^2) = \frac{o^n}{2}$. Based upon this observation, in our experiments we set $o^p = 2$ and $o^n = 0$.

\subsection{Training Objective}
For the Review Helpfulness Prediction problem, the model’s parameters are updated according to the pairwise ranking loss as follows
\begin{equation}
    \mathcal{L}_{\text{ranking}} = \sum_i \text{max} (0, \beta - f(p_i, r^{+}) + f(p_i, r^{-}))
\end{equation}
where $r^{+}$and $r^{-}$ are random reviews in which $r^{+}$ possesses a higher helpfulness level than $r^{-}$. We jointly combine the contrastive goal with the ranking objective of the Review Helpfulness Prediction problem to train our model
\begin{equation}
    \mathcal{L} = \mathcal{L}_\text{AdaptiveCE} + \mathcal{L}_\text{ranking}
\end{equation}
\section{Experiments}
\begin{table}[ht]
\centering
\resizebox{\linewidth}{!}{
\begin{tabular}{l|l|ccc}
\toprule
\multirow{2}{*}{\textbf{Dataset}} & \multirow{2}{*}{\textbf{Split}} & \multicolumn{3}{c}{\textbf{Category (Product / Review)}} \\ 
& & Clothing & Electronics. & Home \\
\midrule
 \multirow{2}{*}{Lazada} & Train \& Dev & 8K/130K & 5K/52K & 4K/16K  \\
  & Test & 2K/32K & 1K/13K & 1K/13K \\
\midrule
 \multirow{2}{*}{Amazon} & Train \& Dev & 16K/349K & 13K/325K & 18K/462K  \\
  & Test & 4K/87K & 3K/80K & 5K/111K \\
 \bottomrule
\end{tabular} }
\caption{
Statistics of MRHP datasets.}
\label{table:datasets}
\end{table}

\begin{table*}[t]
\centering
\resizebox{1\textwidth}{!}{
\begin{tabular}{|l|l|ccc|ccc|ccc|}
\toprule
\multirow{2}{*}{\textbf{Type}} & \multirow{2}{*}{\textbf{Method}} & \multicolumn{3}{c|}{\textbf{Clothing}} & \multicolumn{3}{c|}{\textbf{Electronics}} & \multicolumn{3}{c|}{\textbf{Home}} \\ 
 &  & \textbf{MAP} & \textbf{N@3} & \textbf{N@5} & \textbf{MAP} & \textbf{N@3} & \textbf{N@5} & \textbf{MAP} & \textbf{N@3} & \textbf{N@5} \\
\midrule
\multirow{4}{*}{Text-only} & BiMPM & 60.0 & 52.4 & 57.7 & 74.4 & 67.3 & 72.2 & 70.6 & 64.7 & 69.1 \\
 & EG-CNN & 60.4 & 51.7 & 57.5 & 73.5 & 66.3 & 70.8 & 70.7 & 63.4 & 68.5 \\
 & Conv-KNRM & 62.1 & 54.3 & 59.9 & 74.1 & 67.1 & 71.9 & 71.4 & 65.7 & 70.5 \\
 & PRH-Net & 62.1 & 54.9 & 59.9 & 74.3 & 67.0 & 72.2 & 71.6 & 65.2 & 70.0 \\
\midrule
\multirow{4}{*}{Multimodal} & SSE-Cross & 66.1 & 59.7 & 64.8 & 76.0 & 68.9 & 73.8 & 72.2 & 66.0 & 71.0 \\
 & DR-Net & 66.5 & 60.7 & 65.3 & 76.1 & 69.2 & 74.0 & 72.4 & 66.3 & 71.4 \\
 & MCR & 68.8 & 62.3 & 67.0 & 76.8 & 70.7 & 75.0 & 73.8 & 67.0 & 72.2 \\
 & \textbf{Our Model} & \textbf{70.3} & \textbf{64.7} & \textbf{69.0} & \textbf{78.2} & \textbf{72.4} & \textbf{76.5} & \textbf{75.2} & \textbf{68.8} & \textbf{73.7} \\
\bottomrule
\end{tabular} }
\caption{
Helpfulness Prediction results on Lazada-MRHP dataset.}
\label{table:lazada_results}
\end{table*}

\begin{table*}[t]
\centering
\resizebox{1\textwidth}{!}{
\begin{tabular}{|l|l|ccc|ccc|ccc|}
\toprule
\multirow{2}{*}{\textbf{Type}} & \multirow{2}{*}{\textbf{Method}} & \multicolumn{3}{c|}{\textbf{Clothing}} & \multicolumn{3}{c|}{\textbf{Electronics}} & \multicolumn{3}{c|}{\textbf{Home}} \\ 
 &  & \textbf{MAP} & \textbf{N@3} & \textbf{N@5} & \textbf{MAP} & \textbf{N@3} & \textbf{N@5} & \textbf{MAP} & \textbf{N@3} & \textbf{N@5} \\
\midrule
\multirow{4}{*}{Text-only} & BiMPM & 57.7 & 41.8 & 46.0 & 52.3 & 40.5 & 44.1 & 56.6 & 43.6 & 47.6 \\
 & EG-CNN & 56.4 & 40.6 & 44.7 & 51.5 & 39.4 & 42.1 & 55.3 & 42.4 & 46.7 \\
 & Conv-KNRM & 57.2 & 41.2 & 45.6 & 52.6 & 40.5 & 44.2 & 57.4 & 44.5 & 48.4 \\
 & PRH-Net & 58.3 & 42.2 & 46.5 & 52.4 & 40.1 & 43.9 & 57.1 & 44.3 & 48.1 \\
\midrule
\multirow{4}{*}{Multimodal} & SSE-Cross & 65.0 & 56.0 & 59.1 & 53.7 & 43.8 & 47.2 & 60.8 & 51.0 & 54.0 \\
 & DR-Net & 65.2 & 56.1 & 59.2 & 53.9 & 44.2 & 47.5 & 61.2 & 51.8 & 54.6 \\
 & MCR & 66.4 & 57.3 & 60.2 & 54.4 & 45.0 & 48.1 & 62.6 & 53.5 & 56.6 \\
 & \textbf{Our Model} & \textbf{67.4} & \textbf{58.6} & \textbf{61.6} & \textbf{56.5} & \textbf{47.6} & \textbf{50.8} & \textbf{63.5} & \textbf{54.6} & \textbf{57.8} \\
\bottomrule
\end{tabular} }
\caption{
Helpfulness Prediction results on Amazon-MRHP dataset.}
\label{table:amazon_results}
\end{table*}
\subsection{Datasets}
We evaluate our methods on two publicly available benchmark datasets for MRHP task: Lazada-MRHP and Amazon-MRHP.

\noindent\textbf{Lazada-MRHP} \cite{liu2021multi} consists of product items and artificial reviews on Lazada.com, an e-commerce platform in Southest Asia. All of the texts in the dataset are expressed in Indonesian.

\noindent\textbf{Amazon-MRHP} \cite{liu2021multi} is collected from Amazon.com, the large-scale international e-commerce platform. Product information and associated reviews are in English and extracted between 2016 and 2018.

Both datasets comprise 3 categories: (i) Clothing, Shoes \& Jewelry (Clothing), (ii) Electronics (Electronics), and (iii) Home \& Kitchen (Home). We present the statistics of them in Table \ref{table:datasets}.
\subsection{Implementation Details}
We use a 1-layer LSTM with hidden dimension size of 128. We initialize our word embedding with fastText embedding \cite{bojanowski2017enriching} for Lazada-MRHP dataset and 300-dimensional GloVe pretrained word vectors \cite{pennington2014glove} for Amazon-MRHP dataset. We set our multimodal attention module to have $D = 5$ attention layers. For the visual modality, we extract 2048-dimensional ROI features from each image and encode them into 128-dimensional vectors. Our entire model is trained end-to-end with Adam optimizer \cite{kingma2014adam} and batch size of 32. For the training objective, we set the value of the margin in the ranking loss to be 1.

\subsection{Baselines}
We compare our proposed architecture against the following baselines:
\begin{itemize}
    \item \textbf{BiMPM} \cite{wang2017bilateral}: a ranking model which encodes input sentences in two directions to ascertain the matching result.
    \item \textbf{Conv-KNRM} \cite{dai2018convolutional}: a CNN-based model which encodes n-gram of multiple lengths and uses kernel pooling to generate the final ranking score.
    \item \textbf{EG-CNN} \cite{chen2018cross}: a CNN-based model targeting data scarcity and OOV problem in RHP task via taking advantage of character-based representations and domain discriminators.
    \item \textbf{PRH-Net} \cite{fan2019product}: a baseline to predict helpfulness of a review by taking into consideration both product text and product metadata.
    \item \textbf{DR-Net} \cite{xu2020reasoning}: a cross-modality approach that models contrast in associated contexts by leveraging decomposition and relation modules.
    \item \textbf{SSE-Cross} \cite{abavisani2020multimodal}: multimodal model to fuse different modalities with stochastic shared embeddings.
    \item \textbf{MCR} \cite{liu2021multi}: a baseline model focusing on coherent reasoning.
\end{itemize}

\begin{table*}[ht]
\centering
\resizebox{\textwidth}{!}{
\begin{tabular}{|l|ccc|ccc|ccc|}
\toprule
\multirow{2}{*}{\textbf{Dataset}} & \multicolumn{3}{c|}{\textbf{Clothing}} & \multicolumn{3}{c|}{\textbf{Electronics}} & \multicolumn{3}{c|}{\textbf{Home}} \\ 
 & \textbf{MAP} & \textbf{N@3} & \textbf{N@5} & \textbf{MAP} & \textbf{N@3} & \textbf{N@5} & \textbf{MAP} & \textbf{N@3} & \textbf{N@5} \\
\midrule
 Lazada & $4.48 \cdot 10^{-2}$ & $1.55 \cdot 10^{-2}$ & $3.93 \cdot 10^{-2}$ & $4.54 \cdot 10^{-3}$ & $1.05 \cdot 10^{-4}$ & $2.63 \cdot 10^{-3}$ & $1.09 \cdot 10^{-3}$ & $3.40 \cdot 10^{-2}$ & $3.68 \cdot 10^{-3}$ \\
 Amazon & $3.45 \cdot 10^{-2}$ & $4.22 \cdot 10^{-2}$ & $1.86 \cdot 10^{-2}$ & $4.37 \cdot 10^{-3}$ & $2.81 \cdot 10^{-2}$ & $3.04 \cdot 10^{-2}$ & $2.04 \cdot 10^{-3}$ & $3.30 \cdot 10^{-3}$ & $6.50 \cdot 10^{-3}$ \\
\bottomrule
\end{tabular} }
\caption{
Significance test of the results of our model against MCR model. }
\label{table:sig_tests}
\end{table*}

\subsection{Automatic Evaluation}
In Table \ref{table:lazada_results} and \ref{table:amazon_results}, we follow previous work \cite{liu2021multi} to report Mean Average Precision (MAP), Normalized Discounted Cumulative Gain (NDCG@N) \cite{jarvelin2017ir} where $N = 3$ and $N = 5$. As it can be seen, multimodal approaches achieve better performance than text-only ones.

For Lazada-MRHP dataset, we achieve an absolute improvement of NDCG@3 of 2.4 points in Clothing, NDCG@5 of $1.5$ points in Electronics, and MAP of $1.4$ points in Home over the previous best method, which is MCR. In addition, our model also obtains better results than the best text-only RHP model, which is PRH-Net, with a gain of NDCG@3 of $9.8$ points in Clothing, NDCG@5 of $4.3$ points in Electronics, and MAP of $3.6$ points in Home. Those results prove that our method can produce reasonable rankings for associated reviews.

For Amazon dataset, which is written in English, our model outperforms MCR on all 3 categories, by NDCG@5 of $1.4$ points in Clothing, $2.7$ points in Electronics, and $1.2$ points in Home, respectively. These results have verified that our interaction module and optimization approach can come up with more useful multimodal fusion than previous state-of-the-art baselines, not only in English context but other language one as well. 

We also perform significance tests to evaluate the statistical significance of our improvement on two datasets Amazon-MRHP and Lazada-MRHP, and note p-values in Table \ref{table:sig_tests}. As shown in the table, all of the p-values are smaller than $0.05$, verifying the statistical significance in the enhancement of our method against prior best MRHP model, MCR \cite{liu2021multi}.

\subsection{Case Study}
In Table \ref{table:example}, we introduce an example of one product item and two reviews extracted from Electronics category of Amazon-MRHP dataset. Whereas MCR fails to predict relevant helpfulness scores, our model successfully produces sensible rankings for both of them. We hypothesize that our Multimodal Interaction module learns more meaningful representations and Adaptive Contrastive Learning framework acquires more logical hidden states of relations among input elements. Thus, our model is able to generate more rational outcomes.

\subsection{Ablation Study}
In this section, we proceed to study the impact of (1) Adaptive Contrastive Learning framework and (2) Cross-modal Interaction module.

\noindent\textbf{Adaptive Contrastive Learning} It is worth noting from Table \ref{table:ablation} that plainly integrating contrastive learning brings less enhancement to the performance, with the improvement of NDCG@3 dropping $0.53$ points in Lazada-MRHP dataset, NDCG@5 waning $0.84$ points in Amazon-MRHP dataset. Furthermore, completely removing contrastive objective hurts performance, as NDCG@3 score decreasing $0.77$ points in Lazada-MRHP, and MAP score declining $1.06$ points in Amazon-MRHP. We hypothesize that the model loses the ability to learn efficient representations for cross-modal relations.

\noindent\textbf{Cross-modal Interaction} In this ablation, we eliminate the cross-modal interaction module. As shown in Table \ref{table:ablation}, without the module, the improvement is downgraded, for instance, N@3 drops $1.89$ points in Lazada-MRHP dataset, MAP shrinks $1.39$ points in Amazon-MRHP dataset. It is hypothesized that without the module, the model is rigidly dependent upon the alignment nature among multimodal input elements, which brings about insensible modeling because in most cases, cross-modal elements are irrelevant to be bijectively mapped together.

\begin{table}[ht]
\centering
\resizebox{\linewidth}{!}{
\begin{tabular}{|l|l|ccc|}
\toprule
\textbf{Dataset} & \textbf{Model} & \textbf{MAP} & \textbf{N@3} & \textbf{N@5} \\ 
\midrule
 \multirow{4}{*}{Lazada} & Our Model & \textbf{78.15} & \textbf{72.43} & \textbf{76.49}  \\
  & - w/o Adaptive Weighting & 77.90 & 71.90 & 75.97 \\
  & - w/o Contrastive Objective & 77.69 & 71.66 & 75.85 \\
  & - w/o Cross-modal Module & 77.32 & 70.54 & 74.86 \\
\midrule
 \multirow{4}{*}{Amazon} & Our Model & \textbf{56.49} & \textbf{47.62} & \textbf{50.79}  \\
  & - w/o Adaptive Weighting & 56.03 & 46.98 & 49.95 \\
  & - w/o Contrastive Objective & 55.43 & 46.30 & 49.02 \\
  & - w/o Cross-modal Module & 55.10 & 45.67 & 48.50 \\
 \bottomrule
\end{tabular} }
\caption{
Ablation study in Electronics category of Lazada-MRHP and Amazon-MRHP datasets.}
\label{table:ablation}
\vspace{-10pt}
\end{table}
\subsection{Impact of Contrastive Learning on Cross-modal Relations}
\begin{table*}[t]
\centering
\resizebox{\linewidth}{!}{
\begin{tabular}{|l|l|cc|cc|cc|}
\toprule
\multirow{2}{*}{\textbf{Label}} & \multirow{2}{*}{\textbf{Model}} & \multicolumn{2}{c|}{\textbf{Intra-modal}} & \multicolumn{2}{c|}{\textbf{Inter-modal}} & \multicolumn{2}{c|}{\textbf{Intra-review}} \\ 
 & & \textbf{CS} & \textbf{L2} & \textbf{CS} & \textbf{L2} & \textbf{CS} & \textbf{L2} \\
\midrule
\multirow{2}{*}{1} & MCR & 0.785 $\pm$ 0.002 & 3.852 $\pm$ 0.067 & 0.843 $\pm$ 0.002 & 11.719 $\pm$ 0.001 & 0.845 $\pm$ 0.002 & 14.631 $\pm$ 0.001 \\
 & Our Model & 0.875 $\pm$ 0.002 & 6.545 $\pm$ 0.007 & 0.957 $\pm$ 0.002 & 13.934 $\pm$ 0.027 & 0.953 $\pm$ 0.002 & 15.160 $\pm$ 0.036  \\
 \midrule
 \multirow{2}{*}{4} & MCR & 0.533 $\pm$ 0.004 & 1.014 $\pm$ 0.051 & 0.712 $\pm$ 0.010 & 9.476 $\pm$ 0.001 & 0.617 $\pm$ 0.001 & 8.519 $\pm$ 0.001 \\
 & Our Model & 0.433 $\pm$ 0.001 & 0.981 $\pm$ 0.005 & 0.564 $\pm$ 0.001 & 4.179 $\pm$ 0.017 & 0.538 $\pm$ 0.001 & 3.827 $\pm$ 0.020  \\
\bottomrule
\end{tabular} }
\caption{Intra-modal, Inter-modal, and Intra-review distances in Home category of Lazada-MRHP dataset.}
\label{table:lazada_cs_mse_dist}
\end{table*}

\begin{table*}[t]
\centering
\resizebox{\linewidth}{!}{
\begin{tabular}{|l|l|cc|cc|cc|}
\toprule
\multirow{2}{*}{\textbf{Label}} & \multirow{2}{*}{\textbf{Model}} & \multicolumn{2}{c|}{\textbf{Intra-modal}} & \multicolumn{2}{c|}{\textbf{Inter-modal}} & \multicolumn{2}{c|}{\textbf{Intra-review}} \\ 
 & & \textbf{CS} & \textbf{L2} & \textbf{CS} & \textbf{L2} & \textbf{CS} & \textbf{L2} \\
\midrule
\multirow{2}{*}{1} & MCR & 0.785 $\pm$ 0.006 & 8.532 $\pm$ 0.292 & 0.686 $\pm$ 0.001 & 9.696 $\pm$ 0.300 & 0.880 $\pm$ 0.002 & 9.620 $\pm$ 0.217 \\
 & Our Model & 0.971 $\pm$ 0.001 & 10.663 $\pm$ 0.770 & 0.976 $\pm$ 0.001 & 13.234 $\pm$ 0.493 & 0.970 $\pm$ 0.001 & 12.222 $\pm$ 0.431 \\
 \midrule
 \multirow{2}{*}{4} & MCR & 0.697 $\pm$ 0.009 & 3.045 $\pm$ 0.139 & 0.624 $\pm$ 0.001 & 3.179 $\pm$ 0.830 & 0.781 $\pm$ 0.001	& 5.098 $\pm$ 0.636 \\
 & Our Model & 0.571 +- 0.001 & 1.572 +- 0.037 & 0.488 +- 0.001 & 1.460 +- 0.008 & 0.487 +- 0.001 & 3.555 +- 0.001 \\
\bottomrule
\end{tabular} }
\caption{Intra-modal, Inter-modal, and Intra-review distances in Home category of Amazon-MRHP dataset.}
\label{table:amazon_cs_mse_dist}
\end{table*}

Despite improved performances, it remains a quandary that whether the enhancement stems from more meaningful representations of input samples, which we hypothesize as a significant benefit of our contrastive learning framework. For deeper investigation, we decide to statistically measure distances among input samples using standard distance functions. Table \ref{table:lazada_cs_mse_dist} and \ref{table:amazon_cs_mse_dist} reveal the results of our experiment. In particular, we estimate the cosine distance (CS) and L2 distance (L2) between tokens of (1) product text - review text and product image - review image (intra-modal), (2) product text - review image and product image - review text (inter-modal), and (3) product text - product image and review text - review image (intra-review), then calculate the mean value of all samples. As it can be seen, our frameworks are more efficient in attracting elements of helpful pairs and repelling those of unhelpful pairs.
\section{Related Work}
\subsection{Review Helpfulness Prediction}
Past works that pursue Review Helpfulness Prediction (RHP) dilemma follow text-only approaches. In general, they extract salient information, for instance lexical \cite{krishnamoorthy2015linguistic}, argument \cite{liu2017using}, and emotional features \cite{martin2014prediction} from reviews. Subsequently, these features are fed to a standard classifier such as Random Forest \cite{louppe2014understanding} in order to produce the output score. Inspired by the meteoric development of computation resources, contemporary approaches seek to take advantage of deep learning techniques to tackle the RHP problem. For instance, \citet{wang2017bilateral} propose multi-perspective matching between review and product information via applying attention mechanism. Furthermore, \citet{chen2018cross, dai2018convolutional} adapt CNN models to learn textual representations in various views. 

In reality, review content are not only determined by texts but also other modalities. As a consequence, \citet{fan2019product} integrate metadata information of the target product into the prediction model. \citet{abavisani2020multimodal} filter out uninformative signals before fusing various modalities. Moreover, \citet{liu2021multi} perform coherent reasoning to ascertain the matching level between product and numerous review items. 

\subsection{Contrastive Estimation}
Different from architectural techniques such as Knowledge Distillation \cite{hinton2015distilling, hahn2019self, nguyen2022improving} or Variational AutoEncoder \cite{zhao2020neural, nguyen2021enriching, nguyen2021contrastive, wang2019topic}, Contrastive Learning has been introduced as a representation-based but universal mechanism to enhance natural language processing performance. Proposed by \citet{chopra2005learning}, Contrastive Learning has been widely adopted in myriad problems of Natural Language Processing (NLP). 

As an approach to polish text representations, \citet{gao2021simcse, zhang2021pairwise, liu2021dialoguecse, nguyen2021contrastive} employ contrastive loss to advance sentence embeddings and topic representations. For downstream tasks, \citet{cao2021cliff} propose negative sampling strategies to generate noisy output so that the model can learn to distinguish correct summaries from incorrect ones in Document Summarization. For Spoken Question Answering (SQA), \citet{you2021self} introduce augmentation algorithms in their contrastive learning stage so as to capture noisy-invariant representations of utterances. Additionally, \citet{ke2021classic} inherit the formulation of the contrastive objective to construct distillation loss which transfers knowledge of the previous task to the current one. Their proposals are to improve tasks in the Aspect Sentiment Classification domain. Unfortunately, despite the surge of interest in exercising contrastive learning for NLP, research works to adapt the method to the MRHP task have been scant.
\section{Conclusion}
In this paper, we propose methods to polish representation learning for the Multimodal Review Helpfulness Prediction task.  In particular, we aim to advance cross-modal relation representations by learning mutual information through contrastive learning. In order to further enhance our framework, we propose an adaptive weighting strategy to encourage flexibility in optimization. Moreover, we integrate a cross-modal interaction module to loose the model’s reliance on unalignment nature among modalities, continuing to refine multimodal representations. Our framework is able to outperform prior baselines and achieve state-of-the-art results on the MRHP problem. 
\section{Limitations}
Despite the novelty and benefits of our method for Multimodal Review Helpfulness Prediction (MRHP) problem, it does include some drawbacks. Firstly, even though empirical results demonstrate that our approach not only works in English contexts, we have not conducted the verification in multilingual circumstances, in which product or review texts are written in different languages. If a model is corroborated to work efficiaciously in such contexts, it is capable of providing myriad benefits for practical implementation, for example, e-commerce applications can leverage such one single model for multiple cross-lingual scenarios. Furthermore, our work can also be extended to other domains. For instance, in movie assessment, we need to determine whether the review suits the material in the film, or visual scenes in the comment are consistent with the textual content. These would form our prospective future directions.

Secondly, in the MRHP problem, there are several relationships that contrastive learning could exploit to burnish the performance. In particular, performing contrastive discrimination between two sets of reviews is able to furnish the model with useful set-based representations, which consolidate general knowledge for better helpfulness prediction. Similar insights are applicable for two sets of product information. At the moment, we leave such promising perspectives for future work.
\section{Acknowledgement}
This work was supported by Alibaba Innovative Research (AIR) programme with research grant AN-GC-2021-005.

\bibliography{emnlp2022}

@inproceedings{liu2021multi,
  title={Multi-perspective Coherent Reasoning for Helpfulness Prediction of Multimodal Reviews},
  author={Liu, Junhao and Hai, Zhen and Yang, Min and Bing, Lidong},
  booktitle={Proceedings of the 59th Annual Meeting of the Association for Computational Linguistics and the 11th International Joint Conference on Natural Language Processing (Volume 1: Long Papers)},
  pages={5927--5936},
  year={2021}
}

@inproceedings{zolfaghari2021crossclr,
  title={CrossCLR: Cross-modal Contrastive Learning For Multi-modal Video Representations},
  author={Zolfaghari, Mohammadreza and Zhu, Yi and Gehler, Peter and Brox, Thomas},
  booktitle={Proceedings of the IEEE/CVF International Conference on Computer Vision},
  pages={1450--1459},
  year={2021}
}

@inproceedings{tsai2019multimodal,
  title={Multimodal transformer for unaligned multimodal language sequences},
  author={Tsai, Yao-Hung Hubert and Bai, Shaojie and Liang, Paul Pu and Kolter, J Zico and Morency, Louis-Philippe and Salakhutdinov, Ruslan},
  booktitle={Proceedings of the conference. Association for Computational Linguistics. Meeting},
  volume={2019},
  pages={6558},
  year={2019},
  organization={NIH Public Access}
}

@inproceedings{anderson2018bottom,
  title={Bottom-up and top-down attention for image captioning and visual question answering},
  author={Anderson, Peter and He, Xiaodong and Buehler, Chris and Teney, Damien and Johnson, Mark and Gould, Stephen and Zhang, Lei},
  booktitle={Proceedings of the IEEE Conference on Computer Vision and Pattern Recognition},
  pages={6077--6086},
  year={2018}
}

@inproceedings{vaswani2017attention,
  title={Attention is all you need},
  author={Vaswani, Ashish and Shazeer, Noam and Parmar, Niki and Uszkoreit, Jakob and Jones, Llion and Gomez, Aidan N and Kaiser, {\L}ukasz and Polosukhin, Illia},
  booktitle={Advances in neural information processing systems},
  pages={5998--6008},
  year={2017}
}

@inproceedings{wang2017bilateral,
  title={Bilateral multi-perspective matching for natural language sentences},
  author={Wang, Zhiguo and Hamza, Wael and Florian, Radu},
  booktitle={Proceedings of the 26th International Joint Conference on Artificial Intelligence},
  pages={4144--4150},
  year={2017}
}

@inproceedings{chen2018cross,
  title={Cross-domain review helpfulness prediction based on convolutional neural networks with auxiliary domain discriminators},
  author={Chen, Cen and Yang, Yinfei and Zhou, Jun and Li, Xiaolong and Bao, Forrest},
  booktitle={Proceedings of the 2018 Conference of the North American Chapter of the Association for Computational Linguistics: Human Language Technologies, Volume 2 (Short Papers)},
  pages={602--607},
  year={2018}
}

@inproceedings{dai2018convolutional,
  title={Convolutional neural networks for soft-matching n-grams in ad-hoc search},
  author={Dai, Zhuyun and Xiong, Chenyan and Callan, Jamie and Liu, Zhiyuan},
  booktitle={Proceedings of the eleventh ACM international conference on web search and data mining},
  pages={126--134},
  year={2018}
}

@inproceedings{fan2019product,
  title={Product-aware helpfulness prediction of online reviews},
  author={Fan, Miao and Feng, Chao and Guo, Lin and Sun, Mingming and Li, Ping},
  booktitle={The World Wide Web Conference},
  pages={2715--2721},
  year={2019}
}

@inproceedings{xu2020reasoning,
  title={Reasoning with multimodal sarcastic tweets via modeling cross-modality contrast and semantic association},
  author={Xu, Nan and Zeng, Zhixiong and Mao, Wenji},
  booktitle={Proceedings of the 58th annual meeting of the association for computational linguistics},
  pages={3777--3786},
  year={2020}
}

@inproceedings{abavisani2020multimodal,
  title={Multimodal categorization of crisis events in social media},
  author={Abavisani, Mahdi and Wu, Liwei and Hu, Shengli and Tetreault, Joel and Jaimes, Alejandro},
  booktitle={Proceedings of the IEEE/CVF Conference on Computer Vision and Pattern Recognition},
  pages={14679--14689},
  year={2020}
}

@inproceedings{jarvelin2017ir,
  title={IR evaluation methods for retrieving highly relevant documents},
  author={J{\"a}rvelin, Kalervo and Kek{\"a}l{\"a}inen, Jaana},
  booktitle={ACM SIGIR Forum},
  volume={51},
  number={2},
  pages={243--250},
  year={2017},
  organization={ACM New York, NY, USA}
}

@inproceedings{gao2021simcse,
  title={SimCSE: Simple Contrastive Learning of Sentence Embeddings},
  author={Gao, Tianyu and Yao, Xingcheng and Chen, Danqi},
  booktitle={Proceedings of the 2021 Conference on Empirical Methods in Natural Language Processing},
  pages={6894--6910},
  year={2021}
}

@inproceedings{zhang2021pairwise,
  title={Pairwise Supervised Contrastive Learning of Sentence Representations},
  author={Zhang, Dejiao and Li, Shang-Wen and Xiao, Wei and Zhu, Henghui and Nallapati, Ramesh and Arnold, Andrew O and Xiang, Bing},
  booktitle={Proceedings of the 2021 Conference on Empirical Methods in Natural Language Processing},
  pages={5786--5798},
  year={2021}
}

@inproceedings{liu2021dialoguecse,
  title={DialogueCSE: Dialogue-based Contrastive Learning of Sentence Embeddings},
  author={Liu, Che and Wang, Rui and Liu, Jinghua and Sun, Jian and Huang, Fei and Si, Luo},
  booktitle={Proceedings of the 2021 Conference on Empirical Methods in Natural Language Processing},
  pages={2396--2406},
  year={2021}
}

@inproceedings{cao2021cliff,
  title={CLIFF: Contrastive Learning for Improving Faithfulness and Factuality in Abstractive Summarization},
  author={Cao, Shuyang and Wang, Lu},
  booktitle={Proceedings of the 2021 Conference on Empirical Methods in Natural Language Processing},
  pages={6633--6649},
  year={2021}
}

@inproceedings{you2021self,
  title={Self-supervised Contrastive Cross-Modality Representation Learning for Spoken Question Answering},
  author={You, Chenyu and Chen, Nuo and Zou, Yuexian},
  booktitle={Findings of the Association for Computational Linguistics: EMNLP 2021},
  pages={28--39},
  year={2021}
}

@inproceedings{ke2021classic,
  title={CLASSIC: Continual and Contrastive Learning of Aspect Sentiment Classification Tasks},
  author={Ke, Zixuan and Liu, Bing and Xu, Hu and Shu, Lei},
  booktitle={Proceedings of the 2021 Conference on Empirical Methods in Natural Language Processing},
  pages={6871--6883},
  year={2021}
}

@inproceedings{kim2006automatically,
  title={Automatically assessing review helpfulness},
  author={Kim, Soo-Min and Pantel, Patrick and Chklovski, Timothy and Pennacchiotti, Marco},
  booktitle={Proceedings of the 2006 Conference on empirical methods in natural language processing},
  pages={423--430},
  year={2006}
}

@inproceedings{martin2014prediction,
  title={Prediction of helpful reviews using emotions extraction},
  author={Martin, Lionel and Pu, Pearl},
  booktitle={Proceedings of the AAAI Conference on Artificial Intelligence},
  volume={28},
  number={1},
  year={2014}
}

@article{liu2017using,
  title={Using argument-based features to predict and analyse review helpfulness},
  author={Liu, Haijing and Gao, Yang and Lv, Pin and Li, Mengxue and Geng, Shiqiang and Li, Minglan and Wang, Hao},
  journal={arXiv preprint arXiv:1707.07279},
  year={2017}
}

@article{krishnamoorthy2015linguistic,
  title={Linguistic features for review helpfulness prediction},
  author={Krishnamoorthy, Srikumar},
  journal={Expert Systems with Applications},
  volume={42},
  number={7},
  pages={3751--3759},
  year={2015},
  publisher={Elsevier}
}

@article{alsmadi2020predicting,
  title={Predicting Helpfulness of Online Reviews},
  author={Alsmadi, Abdalraheem and AlZu'bi, Shadi and Al-Ayyoub, Mahmoud and Jararweh, Yaser},
  journal={arXiv preprint arXiv:2008.10129},
  year={2020}
}

@inproceedings{pennington2014glove,
  title={Glove: Global vectors for word representation},
  author={Pennington, Jeffrey and Socher, Richard and Manning, Christopher D},
  booktitle={Proceedings of the 2014 conference on empirical methods in natural language processing (EMNLP)},
  pages={1532--1543},
  year={2014}
}

@article{bojanowski2017enriching,
  title={Enriching word vectors with subword information},
  author={Bojanowski, Piotr and Grave, Edouard and Joulin, Armand and Mikolov, Tomas},
  journal={Transactions of the association for computational linguistics},
  volume={5},
  pages={135--146},
  year={2017},
  publisher={MIT Press}
}

@article{kingma2014adam,
  title={Adam: A method for stochastic optimization},
  author={Kingma, Diederik P and Ba, Jimmy},
  journal={arXiv preprint arXiv:1412.6980},
  year={2014}
}

@inproceedings{chopra2005learning,
  title={Learning a similarity metric discriminatively, with application to face verification},
  author={Chopra, Sumit and Hadsell, Raia and LeCun, Yann},
  booktitle={2005 IEEE Computer Society Conference on Computer Vision and Pattern Recognition (CVPR'05)},
  volume={1},
  pages={539--546},
  year={2005},
  organization={IEEE}
}

@article{louppe2014understanding,
  title={Understanding random forests: From theory to practice},
  author={Louppe, Gilles},
  journal={arXiv preprint arXiv:1407.7502},
  year={2014}
}

@article{nguyen2021contrastive,
  title={Contrastive learning for neural topic model},
  author={Nguyen, Thong and Luu, Anh Tuan},
  journal={Advances in Neural Information Processing Systems},
  volume={34},
  pages={11974--11986},
  year={2021}
}

@inproceedings{hahn2019self,
  title={Self-Knowledge Distillation in Natural Language Processing},
  author={Hahn, Sangchul and Choi, Heeyoul},
  booktitle={Proceedings of the International Conference on Recent Advances in Natural Language Processing (RANLP 2019)},
  pages={423--430},
  year={2019}
}

@inproceedings{nguyen2022improving,
  title={Improving Neural Cross-Lingual Abstractive Summarization via Employing Optimal Transport Distance for Knowledge Distillation},
  author={Nguyen, Thong Thanh and Luu, Anh Tuan},
  booktitle={Proceedings of the AAAI Conference on Artificial Intelligence},
  volume={36},
  number={10},
  pages={11103--11111},
  year={2022}
}

@article{hinton2015distilling,
  title={Distilling the knowledge in a neural network (2015)},
  author={Hinton, Geoffrey and Vinyals, Oriol and Dean, Jeff},
  journal={arXiv preprint arXiv:1503.02531},
  volume={2},
  year={2015}
}

@inproceedings{zhao2020neural,
  title={Neural Topic Model via Optimal Transport},
  author={Zhao, He and Phung, Dinh and Huynh, Viet and Le, Trung and Buntine, Wray},
  booktitle={International Conference on Learning Representations},
  year={2020}
}

@article{nguyen2021enriching,
  title={Enriching and controlling global semantics for text summarization},
  author={Nguyen, Thong and Luu, Anh Tuan and Lu, Truc and Quan, Tho},
  journal={arXiv preprint arXiv:2109.10616},
  year={2021}
}

@inproceedings{wang2019topic,
  title={Topic-Aware Neural Keyphrase Generation for Social Media Language},
  author={Wang, Yue and Li, Jing and Chan, Hou Pong and King, Irwin and Lyu, Michael R and Shi, Shuming},
  booktitle={Proceedings of the 57th Annual Meeting of the Association for Computational Linguistics},
  pages={2516--2526},
  year={2019}
}

@article{han2022sancl,
  title={SANCL: Multimodal Review Helpfulness Prediction with Selective Attention and Natural Contrastive Learning},
  author={Han, Wei and Chen, Hui and Hai, Zhen and Poria, Soujanya and Bing, Lidong},
  journal={arXiv preprint arXiv:2209.05040},
  year={2022}
}
\bibliographystyle{acl_natbib}

\appendix
\newpage
\onecolumn
\section{Hyperspherical Form of Adaptive Contrastive Loss}
We have the initial formulation of the adaptive contrastive loss
\begin{equation}
\mathcal{L}_{\text{AdaptiveCE}} = -\sum_{i=1}^{B} \epsilon^p_i \cdot \text{sim}(\mathbf{t}^1_i, \mathbf{t}^2_i) + \sum_{j=1, k=1, j \neq k}^{B} \epsilon_{j,k}^n \cdot \text{sim}(\mathbf{t}_j^{1}, \mathbf{t}_k^{2})
\end{equation}
We first substitute $\epsilon_i^p = [o^p - \text{sim}(\mathbf{t}^1_i, \mathbf{t}^2_i)]_+$ and $\epsilon_{j,k}^n = [\text{sim}(\mathbf{t}^1_j, \mathbf{t}^2_k) - o^n]_+$ into the above equation,
\begin{align}
& \mathcal{L}_{\text{AdaptiveCE}} = \sum_{i=1}^{B} \text{sim}(\mathbf{t}^1_i, \mathbf{t}^2_i)^2 - o^p \cdot \text{sim}(\mathbf{t}^1_i, \mathbf{t}^2_i)  + \sum_{j=1, k=1, j \neq k}^{B} \text{sim}(\mathbf{t}^1_i, \mathbf{t}^2_i)^2 - o^n \cdot \text{sim}(\mathbf{t}^1_j, \mathbf{t}^2_k) \\
& = \sum_{i=1}^{B} \left(\text{sim}(\mathbf{t}^1_i, \mathbf{t}^2_i) - \frac{o^p}{2}\right)^2 + \sum_{j=1, k=1, j \neq k}^{B} \left(\text{sim}(\mathbf{t}^1_j, \mathbf{t}^2_k) - \frac{o^n}{2}\right)^2 - C 
\end{align}
where $C =  \left(\frac{o^p}{2}\right)^2 + \left(\frac{o^n}{2}\right)^2$. Now we obtain the spherical form of our contrastive loss.

\end{document}